\documentclass[conference]{IEEEtran}
\IEEEoverridecommandlockouts
\usepackage{cite}
\usepackage{amsmath,amssymb,amsfonts}
\usepackage{algorithmic}
\usepackage{graphicx}
\usepackage{textcomp}
\usepackage{booktabs}
\usepackage{multirow} 
\usepackage{url}
\usepackage{comment}
\usepackage{balance}
\usepackage{flushend}
\def\BibTeX{{\rm B\kern-.05em{\sc i\kern-.025em b}\kern-.08em
    T\kern-.1667em\lower.7ex\hbox{E}\kern-.125emX}}
\begin{document}

\title{Facial Forgery-based Deepfake Detection using Fine-Grained Features}

\author{
\IEEEauthorblockN{Aakash Varma Nadimpalli}
\IEEEauthorblockA{\textit{School of Computing} \\
\textit{Wichita State University, Kansas, USA} \\
axnadimpalli@shockers.wichita.edu}
\and
\IEEEauthorblockN{Ajita Rattani}
\IEEEauthorblockA{\textit{Dept. of Computer Science and Engineering} \\
\textit{Uni. of North Texas at Denton, Texas, USA} \\
ajita.rattani@unt.edu}}

\maketitle

\begin{abstract}
Facial forgery by deepfakes has caused major security risks and raised severe societal concerns. As a countermeasure, a number of deepfake detection methods have been proposed. Most of them model deepfake detection as a binary classification problem using a backbone convolutional neural network (CNN) architecture pretrained for the task. These CNN-based methods have demonstrated very high efficacy in deepfake detection with the Area under the Curve (AUC) as high as $0.99$. However, the performance of these methods degrades significantly when evaluated across datasets and deepfake manipulation techniques. This draws our attention towards learning more subtle, local, and discriminative features for deepfake detection. In this paper, we formulate deepfake detection as a fine-grained classification problem and propose a new fine-grained solution to it. Specifically, our method is based on learning subtle and generalizable features by effectively suppressing background noise and learning discriminative features at various scales for deepfake detection. 
Through extensive experimental validation, we demonstrate the superiority of our method over the published research in cross-dataset and cross-manipulation generalization of deepfake detectors for the majority of the experimental scenarios. 
\end{abstract}


\section{Introduction}
\label{sec:intro}
Synthetic media, referred to as deepfakes involve digitally manipulated media (such as face, voice, and text) and pose a significant threat to politics and society. The facial-forgery-based deepfakes specifically denotes the use of deep adversarial models to generate fraudulent content by replacing a person's face with another person's face using manipulation (generation) techniques like FaceSwap and FaceShifter~\cite{Li2019FaceShifterTH}. The utilization of deepfakes for fraudulent activities, evidence falsification, manipulation of public discourse, and disruption of political processes has become a major security concern. To address the risks posed by facial-forgery-based deepfakes, common counter-measures model deepfake detection as a~\textbf{vanilla binary classification} problem~\cite{he2016deep,chollet2017xception,tan2019efficientnet,katamneni2023mis,nadimpalli2023proactive}. Most of them fine-tune convolutional neural networks~(CNN) based classification baselines, such as ResNet-50~\cite{he2016deep}, XceptionNet~\cite{chollet2017xception}, and EfficientNet~\cite{tan2019efficientnet}, for global feature extraction and deepfake detection (real/fake). These CNN-based classification baselines are trained to detect visual artifacts or blending boundaries present in the manipulated images.

However, as the counterfeits become more and more realistic, the differences between real and fake ones will become more subtle and local, thus making such global feature-based vanilla solutions~\textbf{sub-optimal}. Further, these existing global feature-based CNN baselines obtain low generalization in the cross-dataset scenarios and across manipulation techniques~\cite{9857135,afchar2018mesonet,li2020celeb,tan2019efficientnet,Nadimpalli2022GBDFGB,katamneni_nadimpalli_rattani_2023}. The subtle and local property shares a similar spirit as the \textit{fine-grained visual classification problem}. 
Fine-grained classification~\cite{gebru2017fine,he2022transfg} involves categorizing images into very specific and detailed categories, such as different species of birds, dogs, and vehicle models. In contrast to coarse-grained classification, fine-grained classification requires the ability to recognize \textit{subtle differences} in visual features which often exists in small regions referred to as discriminative or foreground regions found using weakly supervised class activation mapping~\cite{hu2019see}, attention mechanism~\cite{8237819}, and vision transformers~(ViT) based on self-attention maps~\cite{he2022transfg}.
A recent study~\cite{chou2023fine} emphasized the importance of suppression of unimportant background regions and fusion of features at various scales for fine-grained visual classification, obtaining state-of-the-art performance in visual classification. 

A handful of facial-forgery-based deepfake detection methods have been proposed based on fine-grained classification~\cite{zhao2021multi,guo2023hierarchical,du2020towards}. Most of these methods are based on texture feature enhancement and the use of an attention mechanism for learning local fine-grained features for deepfake detection.
Inspired by the success of the latest fine-grained classification method~\cite{chou2023fine}, we conjecture that learning subtle features using a fine-grained solution would improve the generalization of the deepfake detectors across datasets and deepfake manipulation techniques. 

To this aim, we propose a novel solution for facial-forgery-based deepfake detection \textbf{based on fine-grained visual classification} that fuses features of varying scales to suppress background noise and learns discriminative features at appropriate scales for an effective and generalizable solution to deepfake detection.

To this aim, the major \textbf{contributions} of this paper are as follows:

\begin{itemize}
    \item A novel method of facial-forgery-based deepfake detection based on learning subtle and generalizable features using a fine-grained solution by effectively fusing features of varying scales suppressing background noise, and learning discriminative features at appropriate scales for the classification. 

    \item Cross-comparison with eight binary classification baselines and published results on cross-dataset~\cite{zhao2021multi,Sun2021DualCL} and cross-manipulation evaluation~\cite{Sun2021DualCL} of deepfake detectors.
   
    
    \item Extensive experiments on publicly available deepfake datasets demonstrating the effectiveness of our method in enhancing the generalization of deepfake detection across datasets and manipulation techniques compared to most of the published work. 
\end{itemize}

This paper is organized as follows: Section~\ref{sec:prior work} discusses the prior work on deepfake detection. Section~\ref{sec:proposed method} describes our proposed deepfake detection based on fine-grained classification. Datasets and implementation details are discussed in section~\ref{sec:Datasets and Experimental Protocol}. Experimental results are discussed in section~\ref{sec:Results and Discussion}. The ablation study is detailed in section~\ref{sec:Ablation study}. Conclusion and future work are discussed in section~\ref{sec:Conclusion and future work}.

\section{Prior Work on DeepFake Detection}
\label{sec:prior work}

In this section, we will discuss the existing countermeasures proposed for facial forgery-based deepfake detection. Most of the existing countermeasures are CNN-based binary classification baselines (based on coarse-grained classification) trained for deepfake detection~\cite{He2016DeepRL,8099678}.

In~\cite{Li_2019_CVPR_Workshops}, VGG16, ResNet50, ResNet101, and ResNet152-based CNNs are fine-tuned for the detection of the presence of artifacts from the facial regions for deepfake detection. Study in~\cite{8630761} proposed two different CNN architectures (Meso-4 and MesoInception-4) composed of only a few layers in order to focus on the mesoscopic properties of the images. Nguyen et al.~\cite{Nguyen2019MultitaskLF} proposed a multi-task CNN to simultaneously detect the fake videos and locate the manipulated regions using an autoencoder with a Y-shaped decoder for information sharing between classification, segmentation, and reconstruction tasks. In~\cite{9717407}, biometric-tailored loss functions (such as Center, ArcFace, and A-Softmax) are used for two-class CNN training for deepfake detection.

Worth mentioning handful of studies in~\cite{zhao2021multi,guo2023hierarchical,du2020towards} have proposed fine-grained classification techniques for facial forgery-based deepfake detection. In~\cite{zhao2021multi}, a new multi-attentional deepfake detection network is proposed that uses multiple spatial attention heads to make the network attend to different local parts. Further aggregation of low-level texture and high-level semantic features guided by attention mechanism are used for deepfake detection. In~\cite{guo2023hierarchical}, the proposed framework consists of three components i.e., multi-branch feature extraction, localization, and classification modules. Each branch of the multi-branch feature extractor learns to classify forgery attributes at one level, while localization and classification modules segment the pixel-level forgery region and detect image-level forgery, respectively. In~\cite{du2020towards} Locality-aware autoEncoder (LAE) is used for feature extraction and subsequent deepfake detection. 

\begin{figure*}[htbp]
\centerline{\includegraphics[width=0.75\textwidth]{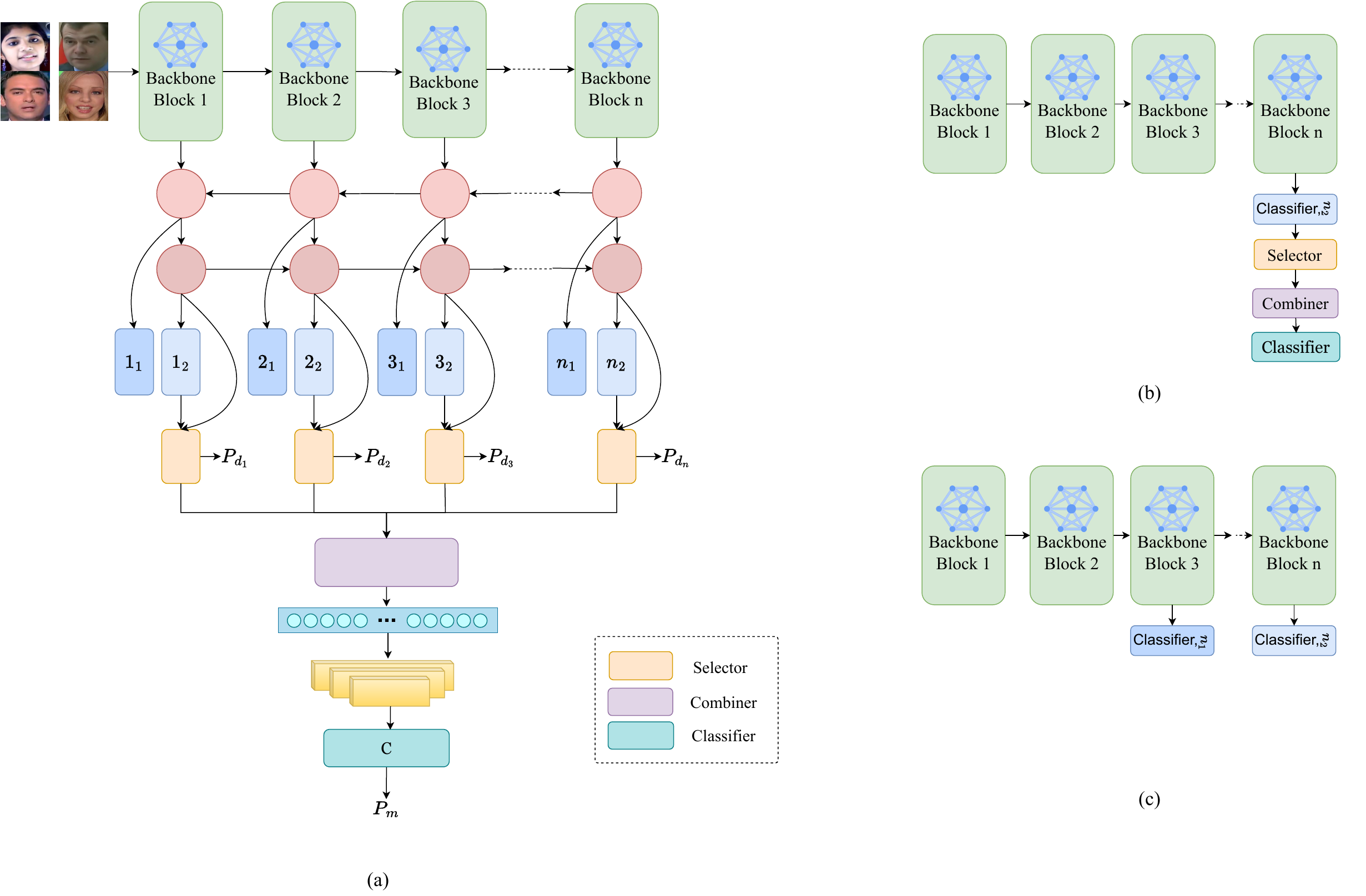}}
\caption{(a) The illustration of the model structure for learning fine-grained features where the initial blocks indicate the backbone blocks of the network, which could be either CNN-based or Transformer-based. The circles in the middle denote the multi-scale feature fusion module, such as the Feature Pyramid Network (FPN) or Path
Aggregation (PA). Then we have classifiers, selectors, and combiner modules for background suppression and multi-scale feature fusion. Features from the combiner are concatenated for the final downstream classification. (b) Illustration of the basic background suppression (BS) module. (c) Illustration of the  basic high-temperature refinement module [best viewed in zoom].}
\label{fig1}
\end{figure*}

\section{Proposed Method}
\label{sec:proposed method}
The proposed network is composed of several components: the backbone, the top-down features-fusion module, the bottom-up feature-fusion module, and finally background suppression (BS) and high-temperature refinement~(refinement) modules. The backbone can be either a convolutional neural network (CNN) or a transformer-based model. The top-down and bottom-up features fusion modules are identical
to the path aggregation network (PA)~\cite{liu2018path} that enhances the receptive field of a network's backbone by aggregating features from different paths within the network.
Figure~\ref{fig1}~(a) shows the overview of the proposed model. The background architecture backbone block is added to the path aggregation (PA) module followed by the classifiers, selectors, and combiner as a part of the background suppression (BS) (Figure~\ref{fig1}~(b)) and high-temperature refinement (refinement) modules (Figure~\ref{fig1}~(c)) detailed below. 

Figure~\ref{fig1} shows the basic illustration of the background suppression (BS) and high-temperature refinement modules, separately. The primary objective of these modules is to learn diverse and discriminative features in order to improve the accuracy of a downstream classification task. The BS module is responsible for suppressing background information and emphasizing the relevant foreground features. The high-temperature refinement module refines the features using a high-temperature mechanism, promoting the exploration of diverse and discriminative representations.

Next, we will discuss the BS and high-temperature refinement modules in detail as follows.

\subsection{Background Suppression (BS)}

The initial step of the background suppression module involves dividing the feature map into two parts, foreground and background, based on classification confidence scores. It then diminishes feature values in areas with low confidence while amplifying distinctive features to enhance their discriminative quality. The background suppression module helps suppress the background noise and irrelevant features.

Let $fs_{i}$ be the feature map associated with the $i^{th}$ backbone block, where $fs_{i}\in R^{C_{i}\times H_{i}\times W_{i}}$. Here, $C_{i}$ represents the number of channels, and $H_{i}$,$W_{i}$ represent the height and width of the feature map respectively. The initial stage of the background suppression (BS) module involves producing classification maps using the feature map. This process can be represented as follows:

\begin{equation}\label{my_first_eqn}
Y_{i}=W_{i}fs_{i}+b_{i}
\end{equation}
               
where $W_{i}$ is the weight of the $i^{th}$ classifier layer, $b_{i}$
is the bias, and $Y_{i}$ is the classification map associated with the classifier layer. Then, finally, a maximum score map is calculated from the classification map which can be expressed as:

\begin{equation}\label{my_second_eqn}
P_{max,i}=max(Softmax(Y_{i}))
\end{equation}

where $P_{max, i}$ represents the $i^{th}$ layer's maximum score map. Next, the features with top-$D_{i}$ scores are selected from all the predictions. The number $D_{i}$ is selected based on the principle that $D_{i} > D_{j}$ when $i<j$. Specifically we set $D_{1}$ to 256, $D_{2}$ to 128, $D_{3}$ to 64, $D_{4}$ to 32. The selection of this value is based on the principle that earlier layers have the potential to  restrict the performance of subsequent layers.

Next, a graph convolution module is utilized to combine the selected features and generate predictions based on the merged features. During this stage, the background suppression (BS) module retains the non-selected classification maps, which are referred to as the dropped maps and it is denoted as $Y_{d}$. Similarly, the merged classification prediction is given by $Y_{m}$. This process is depicted through the selector and combiner components in Figure~\ref{fig1}.

The objective function for the merged classification prediction follows a standard classification approach, utilizing cross-entropy to measure the similarity between the prediction distribution $P_{m}$ and the ground truth label $y$. The merged loss is computed as follows:

\begin{equation}\label{my_third_eqn}
P_{m}=Softmax(Y_{m})
\end{equation}

\begin{equation}\label{my_fourth_eqn}
loss_{m}=-\sum_{c_{i}=1}^{C_{gt}}y_{c_{i}}\log(P_{m,c_{i}})
\end{equation}

Here $y_{c_{i}}$ is the ground truth of $i^{th}$ class, and $P_{m,c_{i}}$ is the predicted probability of the $i^{th}$ class. The summation is conducted across the number of target categories $C_{gt}$. This process amplifies the discriminative features within the selected area.
Another objective of the background suppression (BS) module is to diminish the features in the dropped maps and widen the disparity between the foreground and background. This is accomplished by applying the hyperbolic tangent function (tanh) to the dropped maps, denoted as $Y_d$ (see Eq.~\ref{my_fifth_eqn}, The prediction distribution of this dropped map is given by:

\begin{equation}\label{my_fifth_eqn}
P_{d}=\tanh(Y_{d})
\end{equation}

Subsequently, the dropped loss ($loss_{d}$) is computed as the mean squared error between the prediction and a pseudo target.
\begin{equation}\label{my_sixth_eqn}
loss_{d}=\sum_{c_{i}=1}^{C_{gt}}(P_{d,c_{i}}+1)^{2}
\end{equation}

It is important to note that the hyperbolic tangent function used in Eq.~\ref{my_fifth_eqn} maps the prediction values to a range that is not limited to probabilities. This is because the objective is to effectively distinguish between foreground and background features, even if the background contains appearances of other classes.

To prevent all block feature maps from having high responses only in the same locations, the prediction of each layer is incorporated into the training target in the following manner as follows:

\begin{equation}\label{my_seventh_eqn}
P_{l_{i}}=Softmax(W_{i}(Avgpool(fs_{i}))+b_{i})
\end{equation}
\begin{equation}\label{my_eigth_eqn}
loss_{l}=-\sum_{i=1}^{n}\sum_{c_{i}=1}^{C_{gt}}y_{c_{i}}log(P_{l_{i}},c_{i})
\end{equation}
Here the Avgpool function aggregates all $H_{i}$ and $W_{i}$ values for each channel, where $H_{i}$ represents the height and $W_{i}$  represents the width of the feature map. The number of blocks in the backbone is denoted by $n$.

The overall objective of the background suppression (BS) module is determined by the weighted summation of the merged loss ($loss_m$), dropped loss ($loss_d$), and average layer loss ($loss_l$) as shown in Eq.~\ref{my_ninth_eqn}:
\begin{equation}\label{my_ninth_eqn}
loss_{bs}=\lambda_{m}loss_{m}+\lambda_{d}loss_{d}+\lambda_{l}loss_{l}
\end{equation}
where $\lambda_{m}$, $\lambda_{d}$, and  $\lambda_{l}$ represent the weights assigned to the merged loss, dropped loss, and average layer loss, respectively. In this work, $\lambda_{m}$ is set to $1$,  $\lambda_{d}$ to $5$, and  $\lambda_{l}$ to $0.3$ following the original implementation~\cite{chou2023fine}. These values were selected to strike a balance between the loss incurred by the foreground and background, and they were determined based on the training loss observed during the initial three epochs.

\subsection{High-temperature refinement module (refinement)}

The high-temperature refinement module enables the model to acquire the necessary feature scales by refining the feature map across various scales. This refinement process enhances the learning of diverse features, ultimately improving the model's performance.

For the high-temperature refinement module, classifiers $n_{1}$ and $n_{2}$ w are used followed by the $n^{th}$ block feature map. The classifier $n_{1}$ is positioned within the top-down path, while classifier $n_{2}$ is located in the bottom-up path as shown in Figure~\ref{fig1}. The objective is for classifier $n_{1}$ to learn the output distribution of classifier $n_{2}$. The output of classifier $n_{1}$ is denoted as $Y_{i_{1}}$, while the output of classifier $n_{2}$ is represented as $Y_{i_{2}}$. The refinement objective function facilitates the model in acquiring more diverse and robust representations in the earlier layers while enabling the later layers to concentrate on finer details. This helps in enhancing the overall learning capabilities of the model. To be precise the high-temperature refinement module allows classifier $n_{1}$ to identify broader areas of interest, while classifier $n_{2}$ focuses on learning fine-grained and discriminative features. The refinement loss is computed using the following equations:

\begin{equation}\label{my_tenth_eqn}
P_{i_{1}}=\text{LogSoftmax}(Y_{i_{1}}/T_{e})
\end{equation}

\begin{equation}\label{my_eleventh_eqn}
P_{i_{2}}=\text{Softmax}(Y_{i_{2}}/T_{e})
\end{equation}

\begin{equation}\label{my_fourteenth_eqn}
loss_{r}=P_{i_{2}}\log(\frac{P_{i_{2}}}{P_{i_{1}}})
\end{equation}

where $T_{e}$ represents the temperature at a specific training epoch $e$. The value of $T_{e}$  decreases as the training epoch increases, The $T_{e}$ decay function is defined as:

\begin{equation}\label{my_fifteenth_eqn}
T_{e}=0.5^{[\frac{e}{-\log_{2}(0.0625/T)})]}
\end{equation}

The initial temperature $T$ is set to a relatively high value, such as $64$ or $128$. The objective is to encourage the model to explore a wide range of features, even if the initial predictions may be inaccurate. As the training progresses, the temperature gradually decreases, enabling the model to concentrate more on the target class and learn more discriminative features. This decay policy allows the model to acquire diverse and refined representations, leading to accurate predictions.

The total loss associated with BS and high-temperature refinement modules is defined as:

\begin{equation}\label{my_sixteenth_eqn}
loss_{total}=loss_{bs}+\lambda_{r}loss_{r}
\end{equation}
where $\lambda_{r}$ is the weight for refinement loss, which is set to $1$ following the original implementation~\cite{chou2023fine}.

In this paper, both the background suppression and high-temperature refinement modules have been employed on two different backbone architectures: the Swin transformer-L and EfficientNet-B4. Finally, a \textbf{hybrid model} was developed by concatenating the learned features from the Swin transformer-L and EfficientNet-B4 backbone after applying the proposed fine-grained solution. The concatenated feature vector from Swin transformer-L and EfficientNet-B4 was followed by the dense layer and the final output layer for deepfake detection.

\section{Datasets and Implementation Details}
\label{sec:Datasets and Experimental Protocol}
For all the experiments, we used state-of-the-art publicly available  FaceForensics++~\cite{rossler2019faceforensics++}, Celeb-DF~\cite{li2020celeb} and DFDC~\cite{dolhansky2020deepfake} deepfake datasets. These datasets are discussed as follows:

\begin{itemize}
  \item \textbf{FaceForensics++:} 
  FaceForensics++~\cite{rossler2019faceforensics++} is an automated benchmark for facial manipulation detection. It consists of several manipulated videos created using two different generation techniques: Identity swapping (FaceSwap, FaceSwap-Kowalski, FaceShifter, Deepfakes) and Expression swapping (Face2Face and NeuralTextures). 
 We used the FaceForensics++ dataset's $c23$ version for both training and testing, which has a curated list of $70$ videos for each of these deepfake creation methods.
 \item \textbf{Celeb-DF:} 
 The Celeb-DF~\cite{li2020celeb} deepfake forensic dataset includes $590$ genuine videos from $59$ celebrities as well as $5639$ deepfake videos. Celeb-DF, in contrast to other datasets, has essentially no splicing borders, color mismatch, and inconsistencies in face orientation, among other evident deepfake visual artifacts. The deepfake videos in Celeb-DF are created using an encoder-decoder style model which results in better visual quality. 
  \item \textbf{DeepFake Detection Challenge (DFDC):}
  DFDC (DeepFake Detection Challenge)~\cite{dolhansky2020deepfake} is presently the most extensive face-swapping video dataset accessible to the public. It consists of $1,133$ real videos and $4,080$ fake videos designed specifically for testing forgery detection techniques. The dataset poses significant challenges in deepfake detection due to its diverse nature and the utilization of unknown manipulation methods in creating fake videos. 
\end{itemize}

\noindent \textbf{Implementation Details and Evaluation Metrics:} We implemented the proposed fine-grained solution on Swin Transformer~(Swin-L) and EfficientNet-B4 (CNN) backbone architectures and concatenated them at the feature level (hybrid model) for deepfake detection. These two backbones are chosen due to their proven efficacy in deepfake detection~\cite{liu2021swin,tan2019efficientnet} and other downstream classification tasks. For cross-comparison of our proposed hybrid model, we evaluated eight different deepfake detection baselines based on CNN and Transformer architectures namely ResNet-50~\cite{he2016deep}, XceptionNet~\cite{chollet2017xception}, MesoInceptioNet-4~\cite{afchar2018mesonet}, CNN-LSTM~\cite{de2020deepfake}, EfficientNet-b4~\cite{tan2019efficientnet}, VIT~\cite{dosovitskiy2020image}, Swin-B~\cite{liu2021swin}, and LIT V2-B~\cite{pan2022fast} trained for facial-forgery detection. 

All these models were trained on FF++ $c23$ version which is a high quality (HQ) version of FF++. The face images were detected and aligned using MTCNN. MTCNN utilizes a cascaded CNN-based framework for joint face detection and alignment. The images are then resized to $256\times256$ for both training and evaluation. For all the CNN models, we used a batch-normalization layer followed by the last fully connected layer of size $1024$ and the final output layer for deepfake detection. The CNN models are trained using an Adam optimizer with an initial learning rate of $0.001$, and a weight decay of 1e6. For the Swin Transformer (Swin-L) model, the learning rate is set to $0.0001$, with cosine decay and weight decay set to $0.0003$. The optimizer used is Stochastic Gradient Descend~(SGD). The models are trained on $2$ RTX $8000$Ti GPUs with a batch size of $64$ and number of epochs determined using the early stopping mechanism. These hyper-parameters are determined based on empirical evidence. 
We used the sampling approach described in~\cite{rossler2019faceforensics++} to choose $270$ frames per video for training and $150$ frames per video for validation and testing the models. The trained models are tested on FF++(HQ), DF-$1.0$, Celeb-DF, and DFDC datasets.  

The standard performance metrics used for deepfake detection namely, Area under the Curve~(AUC), Partial Area under the Curve~(pAUC) at $10\%$ False Positive Rate~(FPR), and the Equal Error Rate~(EER) are computed at frame level for the evaluation of the models.

 \begin{table*}
\caption {Evaluation of the existing classification baselines and our proposed hybrid model for deepfake detection in intra- and cross-dataset evaluation. The top performance results are highlighted in bold.}
\label{Table1}
\begin{center}
\scalebox{0.79}{
\begin{tabular}{l|c|lll|lll|lll}
\hline
\multicolumn{1}{c|}{\multirow{2}{*}{Method}} & \multirow{2}{*}{Architecture} & \multicolumn{3}{c|}{FF++}                                                                            & \multicolumn{3}{c|}{Celeb-DF}                                                              & \multicolumn{3}{c}{DFDC}                                                                   \\ \cline{3-11} 
\multicolumn{1}{c|}{}                        &                               & \multicolumn{1}{c|}{AUC}            & \multicolumn{1}{c|}{pAUC}           & \multicolumn{1}{c|}{EER} & \multicolumn{1}{c|}{AUC}            & \multicolumn{1}{l|}{pAUC}           & EER            & \multicolumn{1}{c|}{AUC}            & \multicolumn{1}{l|}{pAUC}           & EER            \\ \hline
ResNet-50                                    & CNN (Global)                  & \multicolumn{1}{l|}{0.945}          & \multicolumn{1}{l|}{0.922}          & 0.125                    & \multicolumn{1}{l|}{0.625}          & \multicolumn{1}{l|}{0.604}          & 0.405          & \multicolumn{1}{l|}{0.621}          & \multicolumn{1}{l|}{0.602}          & 0.409          \\ \hline
XceptionNet                                  & CNN (Global)                  & \multicolumn{1}{l|}{0.985}          & \multicolumn{1}{l|}{0.969}          & 0.037                    & \multicolumn{1}{l|}{0.651}          & \multicolumn{1}{l|}{0.629}          & 0.383          & \multicolumn{1}{l|}{0.649}          & \multicolumn{1}{l|}{0.627}          & 0.385          \\ \hline
MesoInceptionNet-4                           & CNN (Global)                  & \multicolumn{1}{l|}{0.922}          & \multicolumn{1}{l|}{0.898}          & 0.187                    & \multicolumn{1}{l|}{0.598}          & \multicolumn{1}{l|}{0.572}          & 0.459          & \multicolumn{1}{l|}{0.552}          & \multicolumn{1}{l|}{0.529}          & 0.446          \\ \hline
CNN-LSTM                                     & CNN\&LSTM (Global)            & \multicolumn{1}{l|}{0.987}          & \multicolumn{1}{l|}{0.967}          & 0.037                    & \multicolumn{1}{l|}{0.652}          & \multicolumn{1}{l|}{0.631}          & 0.381          & \multicolumn{1}{l|}{0.704}          & \multicolumn{1}{l|}{0.683}          & 0.360          \\ \hline
EN-b4                                        & CNN (Global)                  & \multicolumn{1}{l|}{0.991}          & \multicolumn{1}{l|}{0.975}          & 0.025                    & \multicolumn{1}{l|}{0.688}          & \multicolumn{1}{l|}{0.667}          & 0.369          & \multicolumn{1}{l|}{0.698}          & \multicolumn{1}{l|}{0.676}          & 0.362          \\ \hline
VIT                                          & Transformer (Global)          & \multicolumn{1}{l|}{0.985}          & \multicolumn{1}{l|}{0.966}          & 0.042                    & \multicolumn{1}{l|}{0.698}          & \multicolumn{1}{l|}{0.674}          & 0.362          & \multicolumn{1}{l|}{0.707}          & \multicolumn{1}{l|}{0.685}          & 0.358          \\ \hline
Swin-B                                       & Transformer (Global)          & \multicolumn{1}{l|}{0.988}          & \multicolumn{1}{l|}{0.969}          & 0.034                    & \multicolumn{1}{l|}{0.708}          & \multicolumn{1}{l|}{0.685}          & 0.356          & \multicolumn{1}{l|}{0.714}          & \multicolumn{1}{l|}{0.691}          & 0.354          \\ \hline
LIT V2-B                                     & Transformer (Global)          & \multicolumn{1}{l|}{0.987}          & \multicolumn{1}{l|}{0.965}          & 0.039                    & \multicolumn{1}{l|}{0.719}          & \multicolumn{1}{l|}{0.702}          & 0.349          & \multicolumn{1}{l|}{0.721}          & \multicolumn{1}{l|}{0.698}          & 0.349          \\ \hline
Hybrid model (ours)                         & Fine-Grained                  & \multicolumn{1}{l|}{\textbf{0.995}} & \multicolumn{1}{l|}{\textbf{0.980}} & \textbf{0.018}           & \multicolumn{1}{l|}{\textbf{0.774}} & \multicolumn{1}{l|}{\textbf{0.758}} & \textbf{0.307} & \multicolumn{1}{l|}{\textbf{0.762}} & \multicolumn{1}{l|}{\textbf{0.745}} & \textbf{0.317} \\ \hline
\end{tabular}}
\end{center}
\end{table*}

\begin{table*}
\caption{Evaluation of deepfake detection models trained and tested on the videos from FaceForensics++ dataset across different compression rates. 
The top performance results are highlighted in bold.}
\label{Table2}
\begin{center}
\scalebox{0.80}{
\begin{tabular}{l|lll|lll|lll|lll|lll|lll}
\hline
\multicolumn{1}{c|}{\multirow{2}{*}{Model}} & \multicolumn{3}{c|}{Raw/C23}                                                               & \multicolumn{3}{c|}{Raw/C40}                                                               & \multicolumn{3}{c|}{C23/Raw}                                                               & \multicolumn{3}{c|}{C23/C40}                                                               & \multicolumn{3}{c|}{C40/Raw}                                                               & \multicolumn{3}{c}{C40/C23}                                                                \\ \cline{2-19} 
\multicolumn{1}{c|}{}                       & \multicolumn{1}{l|}{AUC}            & \multicolumn{1}{l|}{pAUC}           & EER            & \multicolumn{1}{l|}{AUC}            & \multicolumn{1}{l|}{pAUC}           & EER            & \multicolumn{1}{l|}{AUC}            & \multicolumn{1}{l|}{pAUC}           & EER            & \multicolumn{1}{l|}{AUC}            & \multicolumn{1}{l|}{pAUC}           & EER            & \multicolumn{1}{l|}{AUC}            & \multicolumn{1}{l|}{pAUC}           & EER            & \multicolumn{1}{l|}{AUC}            & \multicolumn{1}{l|}{pAUC}           & EER            \\ \hline
EN-b4                                       & \multicolumn{1}{l|}{0.981}          & \multicolumn{1}{l|}{0.962}          & 0.046          & \multicolumn{1}{l|}{0.629}          & \multicolumn{1}{l|}{0.614}          & 0.408          & \multicolumn{1}{l|}{0.988}          & \multicolumn{1}{l|}{0.967}          & 0.035          & \multicolumn{1}{l|}{0.686}          & \multicolumn{1}{l|}{0.669}          & 0.352          & \multicolumn{1}{l|}{0.979}          & \multicolumn{1}{l|}{0.960}          & 0.051          & \multicolumn{1}{l|}{0.983}          & \multicolumn{1}{l|}{0.962}          & 0.044          \\ \hline
ViT                                         & \multicolumn{1}{l|}{0.975}          & \multicolumn{1}{l|}{0.957}          & 0.053          & \multicolumn{1}{l|}{0.641}          & \multicolumn{1}{l|}{0.623}          & 0.390          & \multicolumn{1}{l|}{0.979}          & \multicolumn{1}{l|}{0.961}          & 0.048          & \multicolumn{1}{l|}{0.697}          & \multicolumn{1}{l|}{0.681}          & 0.332          & \multicolumn{1}{l|}{0.974}          & \multicolumn{1}{l|}{0.957}          & 0.055          & \multicolumn{1}{l|}{0.981}          & \multicolumn{1}{l|}{0.959}          & 0.048          \\ \hline
Swin-B                                      & \multicolumn{1}{l|}{0.972}          & \multicolumn{1}{l|}{0.954}          & 0.059          & \multicolumn{1}{l|}{0.689}          & \multicolumn{1}{l|}{0.642}          & 0.348          & \multicolumn{1}{l|}{0.985}          & \multicolumn{1}{l|}{0.966}          & 0.040          & \multicolumn{1}{l|}{0.757}          & \multicolumn{1}{l|}{0.742}          & 0.298          & \multicolumn{1}{l|}{0.985}          & \multicolumn{1}{l|}{0.966}          & 0.040          & \multicolumn{1}{l|}{0.978}          & \multicolumn{1}{l|}{0.956}          & 0.051          \\ \hline
Hybrid model (ours)                         & \multicolumn{1}{l|}{\textbf{0.984}} & \multicolumn{1}{l|}{\textbf{0.965}} & \textbf{0.041} & \multicolumn{1}{l|}{\textbf{0.734}} & \multicolumn{1}{l|}{\textbf{0.718}} & \textbf{0.299} & \multicolumn{1}{l|}{\textbf{0.993}} & \multicolumn{1}{l|}{\textbf{0.975}} & \textbf{0.029} & \multicolumn{1}{l|}{\textbf{0.812}} & \multicolumn{1}{l|}{\textbf{0.795}} & \textbf{0.267} & \multicolumn{1}{l|}{\textbf{0.986}} & \multicolumn{1}{l|}{\textbf{0.964}} & \textbf{0.043} & \multicolumn{1}{l|}{\textbf{0.988}} & \multicolumn{1}{l|}{\textbf{0.969}} & \textbf{0.048} \\ \hline
\end{tabular}}
\end{center}
\end{table*}

\begin{table}
\caption{Cross-manipulation evaluation of our hybrid model with published research in~\cite{Sun2021DualCL} in terms of AUC metric. The abbreviations DF, FS, and FST correspond to the DeepFakes, FaceSwap, and FaceShifter-based deepfake manipulation techniques, respectively.}
\label{Table3}
\begin{center}
\scalebox{0.83}{
\begin{tabular}{l|l|l|l|l}
\hline
Train                   & Method                                                                 & DF                                                                            & FS                                                                            & FST                                                                           \\\hline
\multicolumn{1}{c|}{DF} & \begin{tabular}[c]{@{}l@{}}EN-b4~\cite{tan2019efficientnet}\\ MAT~\cite{zhao2021multi}\\ GFF~\cite{Luo2021GeneralizingFF}\\ DCL~\cite{Sun2021DualCL}\\ Hybrid model (ours)\end{tabular} & \begin{tabular}[c]{@{}l@{}}0.997\\ 0.992\\ \textbf{0.998}\\ \textbf{0.998}\\ 0.997\end{tabular} & \begin{tabular}[c]{@{}l@{}}0.462\\ 0.406\\ 0.472\\ 0.610\\ \textbf{0.628}\end{tabular} & \begin{tabular}[c]{@{}l@{}}0.512\\ 0.453\\ 0.519\\ \textbf{0.684}\\ 0.672\end{tabular} \\ \hline
FS                      & \begin{tabular}[c]{@{}l@{}}EN-b4~\cite{tan2019efficientnet}\\ MAT~\cite{zhao2021multi}\\ GFF~\cite{Luo2021GeneralizingFF}\\ DCL~\cite{Sun2021DualCL}\\ Hybrid model (ours)\end{tabular} & \begin{tabular}[c]{@{}l@{}}0.692\\ 0.641\\ 0.702\\ \textbf{0.748}\\ 0.729\end{tabular} & \begin{tabular}[c]{@{}l@{}}0.998\\ 0.996\\ 0.998\\ \textbf{0.999}\\ 0.998\end{tabular} & \begin{tabular}[c]{@{}l@{}}0.607\\ 0.573\\ 0.612\\ 0.648\\ \textbf{0.652}\end{tabular} \\ \hline
FST                     & \begin{tabular}[c]{@{}l@{}}EN-b4~\cite{tan2019efficientnet}\\ MAT~\cite{zhao2021multi}\\ GFF~\cite{Luo2021GeneralizingFF}\\ DCL~\cite{Sun2021DualCL}\\ Hybrid model (ours)\end{tabular} & \begin{tabular}[c]{@{}l@{}}0.611\\ 0.581\\ 0.614\\ 0.639\\ \textbf{0.646}\end{tabular} & \begin{tabular}[c]{@{}l@{}}0.561\\ 0.550\\ 0.561\\ \textbf{0.584}\\ 0.575\end{tabular} & \begin{tabular}[c]{@{}l@{}}\textbf{0.995}\\ 0.991\\ 0.994\\ 0.994\\ 0.993\end{tabular} \\ \hline
\end{tabular}}
\end{center}
\end{table}

\begin{table}
\caption {Comparison of our proposed model with published studies~\cite{zhao2021multi,Sun2021DualCL} on cross-dataset evaluation of deepfake detectors. The results are reported in terms of AUC and EER.}
\label{Table4}
\begin{center}
\scalebox{0.83}{
\begin{tabular}{l|ll|ll|ll}
\hline
\multicolumn{1}{c|}{\multirow{2}{*}{Method}} & \multicolumn{2}{c|}{FF++}                           & \multicolumn{2}{c|}{Celeb-DF}                       & \multicolumn{2}{c}{DFDC}                           \\ \cline{2-7} 
\multicolumn{1}{c|}{}                        & \multicolumn{1}{l|}{AUC}   & EER                    & \multicolumn{1}{l|}{AUC}   & EER                    & \multicolumn{1}{l|}{AUC}   & EER                   \\ \hline
Xception~\cite{chollet2017xception}                                     & \multicolumn{1}{l|}{0.990} & 0.037                  & \multicolumn{1}{l|}{0.651} & 0.383                  & \multicolumn{1}{l|}{0.699} & 0.354                 \\
EN-b4~\cite{tan2019efficientnet}                                        & \multicolumn{1}{l|}{0.992} & 0.036                  & \multicolumn{1}{l|}{0.662} & 0.374                  & \multicolumn{1}{l|}{0.701} & 0.345                 \\
F3-Net~\cite{Qian2020ThinkingIF}                                         & \multicolumn{1}{l|}{0.981} & 0.035                  & \multicolumn{1}{l|}{0.712} & 0.340                  & \multicolumn{1}{l|}{0.728} & 0.333                 \\
Multi-Attention~\cite{zhao2021multi}                                & \multicolumn{1}{l|}{0.993} & \multicolumn{1}{c|}{-} & \multicolumn{1}{l|}{0.674} & \multicolumn{1}{c|}{-} & \multicolumn{1}{c|}{-}     & \multicolumn{1}{c}{-} \\
GFF~\cite{Luo2021GeneralizingFF}                                            & \multicolumn{1}{l|}{0.983} & 0.038                  & \multicolumn{1}{l|}{0.753} & 0.324                  & \multicolumn{1}{l|}{0.715} & 0.347                 \\
Cross-Modality ~\cite{DBLP:journals/nca/ZhaoZDC23}                       & \multicolumn{1}{l|}{\textbf{0.998}} & \multicolumn{1}{c|}{-} & \multicolumn{1}{l|}{0.769} & \multicolumn{1}{c|}{-} & \multicolumn{1}{l|}{0.790} & \multicolumn{1}{c}{-} \\
MA Localization~\cite{syed2022multi}                              & \multicolumn{1}{l|}{0.957} & \multicolumn{1}{c|}{-} & \multicolumn{1}{l|}{0.672} & \multicolumn{1}{c|}{-} & \multicolumn{1}{l|}{\textbf{0.791}} & \multicolumn{1}{c}{-} \\
LTW~\cite{Sun2021DomainGF}                                            & \multicolumn{1}{l|}{0.991} & 0.033                  & \multicolumn{1}{l|}{0.771} & 0.293                  & \multicolumn{1}{l|}{0.745} & 0.338                 \\
DCL~\cite{Sun2021DualCL}                                            & \multicolumn{1}{l|}{0.993} & 0.032                  & \multicolumn{1}{l|}{\textbf{0.823}} & \textbf{0.265}                  & \multicolumn{1}{l|}{0.767} & {0.319}                \\\hline

Hybrid model (ours)                          & \multicolumn{1}{l|}{0.995} & \textbf{0.030}                  & \multicolumn{1}{l|}{0.774} & 0.297                  & \multicolumn{1}{l|}{0.762} & \textbf{0.317}                 \\ \hline
\end{tabular}}
\end{center}
\vspace{-6mm}
\end{table}

\begin{table}
\caption {Evaluation of our model in terms of AUC when different modules are added to the backbone networks. }

\label{Table5}
\begin{center}
\scalebox{0.85}{
\begin{tabular}{ll|cc}
\hline
\multicolumn{2}{c|}{Module}                      & \multicolumn{2}{c}{Hybrid Model}                                           \\ \hline
\multicolumn{1}{l|}{Refinement}   & BS           & \multicolumn{1}{c|}{Celeb-DF}       & \multicolumn{1}{l}{FF++(C23)} \\ \hline
\multicolumn{1}{l|}{}             &              & \multicolumn{1}{c|}{0.744}          & 0.989                         \\ \hline
\multicolumn{1}{l|}{$\checkmark$} &              & \multicolumn{1}{c|}{0.762}          & 0.994                         \\ \hline
\multicolumn{1}{l|}{}             & $\checkmark$ & \multicolumn{1}{c|}{0.755}          & 0.991                         \\ \hline
\multicolumn{1}{l|}{$\checkmark$} & $\checkmark$ & \multicolumn{1}{c|}{\textbf{0.774}} & \textbf{0.995}                \\ \hline
\end{tabular}}
\vspace{-4mm}
\end{center}
\end{table}


\section{Results and Discussion}
\label{sec:Results and Discussion}
In this section, we will discuss the results of our proposed model across datasets, compression rates, and deepfake manipulation (generation) techniques. 

\subsection{Intra- and cross-dataset evaluation with binary baselines} Table~\ref{Table1} shows the performance of all the baseline classification baselines (both CNN and Transformer-based) and our proposed model (hybrid model) in the intra- and cross-dataset scenarios for facial forgery-based deepfake detection. All the models are trained on FaceForensics++~(C23) and tested on FaceForensics++~(C23), DFDC and Celeb-DF datasets. The performances are reported in terms of AUC, pAUC, and EER. 

The top performance results are highlighted in bold across various evaluation datasets. It can be seen that all the models obtained high performance in the intra-dataset evaluation scenario i.e. when the training and test sets are 
FaceForensics++~(C23). However, the performance drop of the models across datasets is significant for both the Celeb-DF and DFDC datasets. The reason is that deepfake videos in Celeb-DF are created using an encoder-decoder style model which results in better visual quality. Similarly, the fake videos in DFDC datasets are more diverse and have unknown manipulation methods. Therefore, all the models have obtained performance degradation due to domain shift i.e., change in the training and testing data distribution.   

However, worth noting that our proposed model outperforms all the baseline models in the intra- as well as the cross-dataset scenario (see Table~\ref{Table1}). Our proposed model obtained the best results with an overall performance increment of $0.004$, and $0.005$ in terms of AUC, pAUC, and EER reduction of $0.007$, when compared to second best model (i.e., EN-b4) when trained and evaluated on FF++ dataset. Similarly, on the Celeb-DF dataset, our proposed model obtained the best results with an overall performance increment of $0.055$ and $0.056$ in terms of AUC and pAUC, and EER reduction of $0.042$ when compared to the second-best model (i.e., LIT V2-B). Finally, on the DFDC dataset, our proposed model also obtained the best results with an overall performance increment of $0.041$ and $0.047$ in terms of AUC and pAUC and EER reduction of $0.032$, when compared to second-best model (i.e, LIT V2-B). 

Thus, these experimental results demonstrate the efficacy of our proposed model in learning efficient features that are generalizable across datasets in comparison to vanilla-based classification baselines.

\begin{figure*}[htbp]
\centerline{\includegraphics[width=0.50\textwidth]{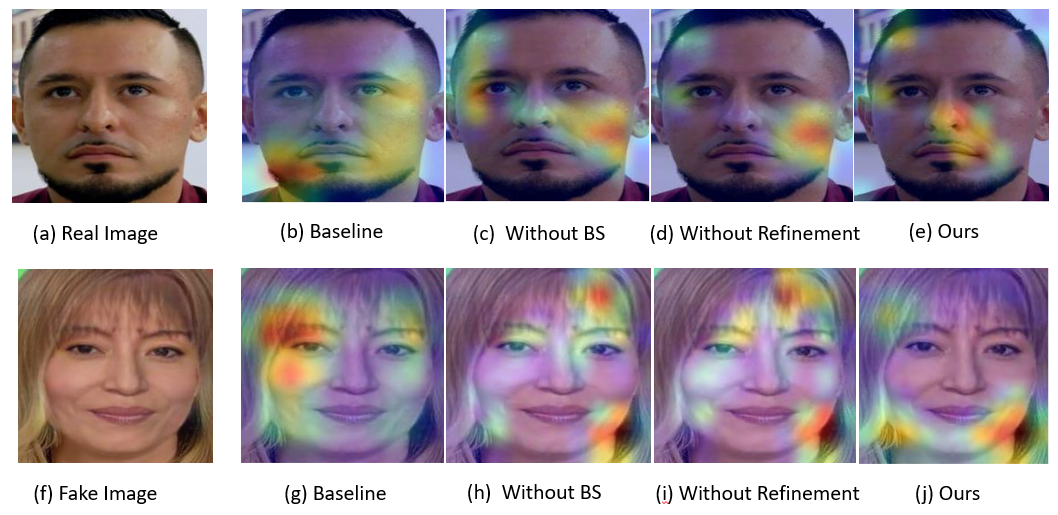}}
\caption{Grad-CAM visualization of our proposed hybrid model with baseline, and without BS module and Refinement modules.}
\label{fig2}
\vspace{-3mm}
\end{figure*}

\subsection{Evaluation across different video compression rates}
Existing studies~\cite{wu2022ggvit,zhao2021multi} suggest that the performance of the deepfake detector degrades on the compressed videos. This is due to the loss of high-frequency components, as a result of compression, which represents the introduced artifacts in fake videos used for deepfake detection. Therefore, we also evaluated the efficacy of the proposed model on videos compressed at different compression rates as a part of FaceForensics++ dataset. Cross-comparison of our proposed model with the four best-performing existing baselines is also shown in Table~\ref{Table2}.

As can be seen in Table~\ref{Table2}, our proposed model significantly enhances generalization on the samples across compression rates. In this Table, $C$ implies the compression rate. For instance, Raw/C40 indicates that the training dataset has raw (C0) i.e., no compression, and the testing dataset is compressed at $40\%$. 
The best results in terms of AUC, pAUC, and EER are obtained by our proposed model on the C23/Raw setting with an overall performance increment of $0.005$, and $0.008$ in terms of AUC, and pAUC, and EER reduction of $0.006$, when compared to second best model (i.e., EN-b4). The lowest performance in terms of AUC, pAUC, and EER is obtained by the EN-B4 model on Raw/C40 setting with an overall performance decrement of $0.105$, and $0.104$ in terms of AUC, pAUC, and EER increment of $0.109$, when compared to our best model (i.e., Hybrid Model). Our proposed hybrid model outperformed all the other models at different compression rates.

\subsection{Cross manipulation evaluation}
To further our investigation, we carried out experiments across different deepfake manipulation techniques to delve deeper into the model's capacity to generalize across unknown deepfake manipulations. 
We specifically selected the DeepFakes (DF), FaceSwap (FS), and FaceShifter (FST) based manipulation techniques from the FF++(C23) dataset. All of these are different manipulation techniques for face-swapping-based deepfake generation. For the purpose of this experiment, all the models are trained on one of these manipulation techniques and subsequently evaluated on the other two, enabling us to assess their generalizability across different manipulation techniques. 
In contrast to FaceShifter (FST), which permits unrestricted face swapping for a single face image, DeepFakes (DF) needs training using pairs of swapped faces. Thus the approach to manipulation used by these two techniques diverges significantly. 

Table~\ref{Table3} shows the performance of our hybrid model over existing models across facial manipulation techniques. The results of the evaluation of the existing deepfake models across manipulation techniques are taken from~\cite{Sun2021DualCL}. As can be seen, our method consistently performs better than the other models in terms of the AUC by $0.01$ when dealing with unseen manipulation types. The Dual contrastive learning (DCL) model~\cite{Sun2021DualCL} obtained better results on some manipulation types but overall a comparable performance is obtained with our proposed model.

Given the disparities in the manipulation techniques, the effectiveness of cross-manipulation approaches depends on the robustness of the learned feature representation spotting the discriminative regions between real and synthetic faces. 
Our proposed model efficiently distinguishes between real and fake faces across different manipulation techniques. 

These experiments suggest the efficacy of our proposed model in learning fine-grained features for deepfake detection that are robust across datasets and manipulation techniques over existing binary classification baselines.

\subsection{Comparison with published results on cross-dataset evaluation} In Table~\ref{Table4}, we compared the performance of our hybrid model with published results on cross-dataset evaluation of deepfake detectors~\cite{zhao2021multi,Sun2021DualCL}. All the deepfake detectors are trained on FaceForensics++ (FF++) high-quality version and tested on FaceForensics++(HQ), Celeb-DF, and DFDC datasets. 

Our proposed model obtained \emph{equivalent performance} with most of the best-performing methods in intra-dataset evaluation. At the same time, our proposed model \emph{outperformed most of the existing methods in cross-dataset evaluation} on Celeb-DF and DFDC datasets. On the Celeb-DF dataset, Our proposed model obtained the second-best results with an overall performance decrement of only $0.049$ in terms of AUC when compared to the best model (i.e., DCL). Similarly, On the DFDC dataset, our proposed model obtained the best results with an overall EER reduction of $0.002$ when compared to the second-best model (i.e., DCL). Thus, obtaining state-of-the-art performance. Recall that due to the difference in the deepfake generation techniques between FaceForensics++(HQ), Celeb-DF, and DFDC datasets, the performance drop of all the models in Table~\ref{Table4} is significant on cross-dataset evaluation.

Although, DCL~\cite{Sun2021DualCL}  obtained better results with an AUC of $0.823$ over our model ($0.774$) in cross-dataset evaluation on the Celeb-DF.
However, our model obtained a lower Equal error rate~(EER) of $0.317$ when compared to the EER of the DCL model ($0.319$) when tested on the DFDC dataset. The DCL method uses a data-view generation module to generate different views of input samples by specially designed data augmentation, and then constructs positive and negative data pairs and performs contrastive learning at different granularities to improve generalization across the dataset. Despite the enhanced performance, DCL primarily relies on the generation of paired images, which is usually unpredictable in practice. 
Similarly, the MA Localization model~\cite{syed2022multi} (fine-grained deepfake detector) obtained better results than our model with an overall performance increment of $0.029$ in terms of AUC on DFDC dataset. However, our model outperformed this model  with an overall performance increment of $0.102$ in terms of AUC on the Celeb-DF dataset.

These results confirm that our hybrid fine-grained model surpasses the majority of current approaches to facial-forgery-based deepfake detection, demonstrating robust performance across datasets, compression rates, and manipulation techniques.

\section{Ablation study}
\label{sec:Ablation study}
In this section, we did an Ablation study to better understand the impact of each module used in our proposed model. To this front, we separately added Refinement and Background suppression (BS) modules to the classification backbones by keeping the path aggregation (PA) module constant. 

As a baseline, we trained the Swin transformer Large (Swin-L) and Efficient Net-B4 model on FF++ without the refinement and BS modules. We used the concatenated deep features from these trained models for deepfake detection.
Further, we trained the Swin transformer Large (Swin-L) and Efficient Net-B4 model on FF++ without the refinement and BS modules, alternatively, and concatenated their features for the final deepfake detection. 

Table~\ref{Table5} shows the complete configuration of this Ablation Study and the results obtained on FF++ and Celeb-DF datasets. 
The hybrid model with refinement and BS modules has the best performance with an overall increment of $0.012$, and $0.001$ in terms of AUC on Celeb-DF and FF++ datasets, respectively, when compared to our model with only the refinement module. This suggests the importance of suppression of background or unimportant regions for fine-grained classification in line with~\cite{chou2023fine}. 

Finally, we also used Explainable AI (XAI) based Gradient weighted Class
Activation Mapping (Grad-CAM)~\cite{8237336} visualization to understand the distinctive image regions used by our model with different module settings. Figure~\ref{fig2} shows the Grad-CAM~\cite{8237336} visualization of our proposed hybrid model on samples from the FF++ dataset with baseline (hybrid combination of Swin-L and EfficientNet-B4 without fine-grained solution), Without BS module, Without Refinement module, and Ours (hybrid model with both BS and Refinement modules). It can be seen that the model gradually concentrates on the finer details when both modules are included with the backbone architectures. This confirms the importance of the joint use of background suppression (BS) and refinement modules for noise removal and discriminative feature learning for fine-grained deepfake detection. 

\section{Conclusion and future work}
\label{sec:Conclusion and future work}

In this paper, we improve the cross-dataset and cross-manipulation generalization of the deepfake detectors by proposing a fine-grained solution to it. 
Our proposed facial-forgery-based deepfake detection is based on learning subtle and generalizable features by effectively fusing features of varying scales suppressing background noise, and learning discriminative features at appropriate scales for the classification. 
Experimental results demonstrate the merit of our fine-grained solution in improving the cross-dataset and cross-manipulation generalization performance of the deepfake detector over the existing classification baselines for most of the experimental scenarios. The proposed model could be used with any backbone model for deepfake detection. As a part of future work, experimental evaluations will be extended on different backbone architectures and datasets with transformations, such as cropping and adversarial noise, to draw further insights into our fine-grained solution to deepfake detection.

\small{
\flushend
\bibliographystyle{IEEEtran}
\bibliography{Main}}
\end{document}